%% file: naacl2021.tex
\pdfoutput=1
\documentclass[11pt]{article}

\usepackage{naacl2021}

\usepackage{times}
\usepackage{latexsym}

\usepackage[T1]{fontenc}
\usepackage[utf8]{inputenc}

\usepackage{microtype}
\usepackage{ifthen}
\usepackage{color}
\usepackage[utf8]{inputenc}
\usepackage{url}
\usepackage{multicol}
\usepackage{tikz}
\usepackage{booktabs}
\usepackage{csquotes}
\usepackage{soul}
\usepackage{float}
\usetikzlibrary{fit}
\usetikzlibrary{patterns}

\usepackage{enumitem}
\usepackage{tabularx}

\usepackage{booktabs}
\usepackage{amssymb}
\usepackage{xspace}

\usepackage{bbm}
\usepackage{amsmath}
\usepackage{url}
\usepackage{comment}
\usepackage{subcaption}

\usepackage{url}
\usepackage{enumitem}
\usepackage{booktabs}
\usepackage{multirow}
\usepackage{graphicx}
\usepackage{array}
\usepackage{subcaption}
\usepackage{amsthm,amssymb,amsopn,bm}
\usepackage{color,soul}
\usepackage{verbatim}
\usepackage{caption}
\usepackage{xcolor,colortbl}
\definecolor{green}{rgb}{0.1,0.1,0.1}
\usepackage{tabulary}
\usepackage{footnote}
\usepackage{blindtext}
\usepackage{algorithm}
\usepackage[noend]{algpseudocode}
\usepackage{stmaryrd}
\usepackage{CJKutf8}
\usepackage[OT2,T1]{fontenc}
\usepackage[russian, english]{babel}
\usepackage{multirow}
\makesavenoteenv{tabular}
\makesavenoteenv{algorithm}
\usepackage{tabularx,booktabs}
\newcolumntype{Y}{>{\centering\arraybackslash}X}

\setlist{leftmargin=8mm}

\usepackage{enumitem}

\usepackage{tikz}

\definecolor{gitred}{HTML}{FDB8C0}
\definecolor{gitgreen}{HTML}{006400}
\definecolor{chocolate}{HTML}{D2691E}
\definecolor{maroon}{HTML}{800000}
\definecolor{indigo}{HTML}{4B0082}
\definecolor{green}{HTML}{008000}
\definecolor{orange}{HTML}{fc8d62}
\definecolor{purple}{HTML}{8da0cb}
\title{
Capturing Label Distribution: 
A Case Study in NLI}

\author{Shujian Zhang  \qquad Chengyue Gong \qquad  Eunsol Choi \\
Department of Computer Science, The University of Texas at Austin }
\begin{document}
\maketitle

\begin{abstract}%by exploring calibration methods and introducing diverse labels per example into training
We study estimating inherent human disagreement (annotation label distribution) in natural language inference task. Post-hoc smoothing of the predicted label distribution to match the expected label entropy is very effective. Such simple manipulation can reduce KL divergence by almost half, yet will not improve majority label prediction accuracy or learn label distributions. To this end, we introduce a small amount of examples with multiple references into training. We depart from the standard practice of collecting a single reference per each training example, and find that collecting multiple references can achieve better accuracy under the fixed annotation budget. Lastly, we provide rich analyses comparing these two methods for improving label distribution estimation.% and to intepret distributional metrics. %the gains from calibration methods and multi-annotated training. %Overall, our analysis suggests distribution metrics should be interpreted more carefully.  %We offer rich analysis of label entropy distribution and % can improve not only the label distribution but also accuracy. Lastly, we showing that evaluation of distribution should be studied more carefully. %While most existing data collection efforts focuses on covering diverse training examples, weWe also study a recently-proposed setting that captures human disagreement, where model is tasked with predicting the label distribution instead of a single label. We find this approach of collecting diverse training instances shows even bigger gain on predicting a better label distribution.  inherent human disagreement in 
\end{abstract}

\input{tex/intro.tex}

\input{tex/evaluation.tex}
\input{tex/related.tex}
\input{tex/conclusion}
% \newpage
% \input{tex/ethics}

\section*{Acknowledgements}
The authors thank Greg Durrett, Raymond Mooney, and Michael Zhang for helpful comments on the paper draft. 

\bibliography{custom}
\bibliographystyle{acl_natbib}

\clearpage
\onecolumn
\appendix
\input{tex/appendix}

% \section{Example Appendix}
% \label{sec:appendix}

\end{document}

%% file: tex/intro.tex
\section{Introduction}\label{sec:intro}

Recent papers~\cite{Pavlick2019InherentDI,Nie2020WhatCW} have shown that human annotators disagree on solving natural language inference (NLI) tasks~\cite{Dagan2005ThePR,Bowman2015ALA}, which decides whether hypothesis \textit{h} is true given premise \textit{p}. Such disagreement is not an annotation artifact but rather exhibits the judgement of annotators with differing interpretations of entailment~\cite{Reidsma2008ExploitingA}. We study how to estimate such \textbf{distribution} of labels for NLI task, as introduced in newly proposed evaluation dataset~\cite{Nie2020WhatCW} which contains 100 human labels per example.%, and their task of predicting the distribution over labels instead of label classification. 

%More specifically, the task is casted as a classification over three options: entailment, contradiction and neutral. 

Without changing model architectures~\cite{bert}, we focus on improving predicted label distribution. We introduce two simple ideas: calibration and training with examples with multiple references ({multi-annotated}). Estimating label distribution is closely related to calibration~\cite{Raftery2005UsingBM}, which studies how aligned is the predicted probability distribution with empirical likelihood, in this case, human label distribution. When trained naively with cross entropy loss on unambiguously annotated data (examples with a single label), models generate a over-confident distribution~\cite{Thulasidasan2019OnMT} putting a strong weight on a single label. Observing that the entropy of predicted distribution is substantially lower than that of human label distribution, we calibrate this distribution such that two distributions have comparable entropy via temperature scaling~\cite{Guo2017OnCO} and label smoothing \citep{szegedy2016rethinking}.
%Observing that our models tend to be over confident~\cite{} after trained exclusively , we use temperature scaling and label smoothing change the prediction distribution. our focus on improving the distribution  method improves XX by XX and XX by XX. We introduce two ideas to show improvements: distribution calibration and more efficient label annotation distribution. , which is increasingly important as NLP techniques are deployed in high-stake applications where interpretability is a key
%This simple langauge inference tasks has been widely adopted benchmark~\cite{Bowman2015ALA,Williams2018ABC} in the NLP community. 

%We discover that measuring model performance with Jensen-Shannon Distance (JSD) and the Kullback–Leibler (KL) divergence between model softmax outputs and the estimated distribution over human annotation should be done carefully, i.e., after correctly smooth model distribution. 

%Throughout the training, model is exposed to only a single label for each example. incentivizing models to predict distribution. W

%This simple approach shows significant gains, reducing 
While such calibration shows strong gains, reducing the KL divergence~\cite{kullback1951information} between predicted and human distribution by roughly half, it does not improve accuracy. Prior works introduce inherent human disagreement into evaluation, and we further embrace such ambiguity into \textit{training}. %Recent NLP benchmark datasets~\citep[e.g.][]{Bowman2015ALA, Rajpurkar2016SQuAD10} collect multiple references for the evaluation set. However, 
Almost all nlp datasets~\cite{Wang2019SuperGLUEAS,Rajpurkar2016SQuAD10} present single reference for training examples while collecting multiple references for examples in the evaluation dataset. We show that, under the same annotation budget, adding a small amount of multi-annotated training, at the cost of decreasing total examples annotated, improves label distribution estimation as well as prediction accuracy. Lastly, we provide rich analyses on differences between calibration approach and multi-annotation training approach. To summarize, our contributions are the following:

 \begin{itemize}
   \item Introduce calibration techniques to improve label distribution estimation in natural language inference task.
   \item Present an empirical study showing collecting multiple references for a small number of training examples is more effective than labeling as many examples as possible, under the same annotation budget. 
   \item Study the pitfalls of using distributional metrics to evaluate human label distribution.
%   Propose that NLI evaluation should incentivize models over human judgements instead of to predict a single label into training. 
 \end{itemize}

%% file: tex/evaluation.tex
\section{Evaluation}
\begin{table*}
\centering
\footnotesize
    \centering
    \begin{tabular}{l|l|r|r|r|r|r|r|r}
\toprule
 &   \multicolumn{4}{c}{ChaosSNLI ($H=0.563$)} & \multicolumn{4}{|c}{ChaosMNLI ($H=0.732$)}  \\ 
   & JSD$\downarrow$  & KL $\downarrow$ &  acc (old/new)$\uparrow$  & $H$ & JSD $\downarrow$ & KL $\downarrow$ &  acc (old/new) $\uparrow$  & $H$  \\ \midrule
%Chance (all)    & 0.383 & 0.545 & 0.447 / 0.537 && 0.302 & 0.3559 & 0.451 / 463 & 0.3205 \\%& 0.406 & 0.509 / 0.502& \\
%Chance   &  \\
\textcolor{gray}{Best reported (all))} & 0.220 & 0.468 & 0.749 / 0.787 &&0.305 &0.665 &0.674 / 0.635 & \\
\textcolor{gray}{Est. human} & 0.061 & 0.041 & 0.775 / 0.940 & & 0.069 & 0.038 & 0.660 / 0.860 & \\
\textcolor{gray}{RoBERTa (all)}   & 0.229 & 0.505& 0.727 / 0.754 &  & 0.307 & 0.781& 0.639 / 0.592  \\ \midrule 
\textcolor{gray}{RoBERTa (our reimpl.,all)} & 0.230 & 0.502& 0.723 / 0.754 &  & 0.310 & 0.790& 0.642 / 0.594 \\
RoBERTa (our reimpl.,subset) & 0.242&0.548&	0.684 / 0.710&	0.345 & 0.308 & 0.799 & 0.670 / 0.604 & 0.414 \\
+ calib (temp. scaling) & 0.202 &0.281&	0.684 / 0.710&	0.569 & 0.233 & 0.324 & 0.670 / 0.604 & 0.720 \\
+ calib (pred smoothing)& 0.222	&0.326&	0.684 / 0.710&	0.566 & 0.245 & 0.347 & 0.670 / 0.604 & 0.722\\ % $\alpha= 0.1$ & & &0.222& 	0.344&	0.684 / 0.710&	0.529 & 0.267 & 0.462 & 0.670/ 0.604 & 0.579& \\$\alpha= 0.2$ \
+ calib (train smoothing)& 0.221& 0.338& 0.688 / 0.710 &0.537 & 0.252 & 0.372 & \textbf{0.680} / 0.602 & 0.701\\
+ multi-annot &\textbf{0.183}&	0.203&	\textbf{0.690 / 0.740}&	0.649  & \textbf{0.190} & \textbf{0.179} & 0.646 / \textbf{0.690} & 0.889 \\ \
+ pred smoothing \& multi-annot & 0.202&	\textbf{0.196}&	\textbf{0.690 / 0.740}&	0.773  & 0.209 & 0.189 & 0.646 / \textbf{0.690} & 0.977   \\ \bottomrule
    \end{tabular}
    \caption{Main results: $H$ next to the dataset name on the top row refers to the entropy value of human label distribution. All calibration methods show significant gains on distribution metrics, but introducing multi-annotated examples into training shows the strongest results. The top block results are from ~\citet{Nie2020WhatCW}, and rows in grey color are not strictly comparable (evaluated on the different set).}
    \label{tab:main_result}
\end{table*}

\subsection{Data}
We use the training data from the original SNLI~\cite{Bowman2015ALA} and MNLI dataset~\cite{Williams2018ABC}, each containing 592K and 392K instances. We evaluate on ChaosNLI dataset~\cite{Nie2020WhatCW}, which is collected the same way as original SNLI dataset, but contains 100 labels per example instead of five.\footnote{It covers SNLI, MNLI, and $\alpha$NLI~\cite{bhagavatula2020abductive}, and we focus our study on the first two datasets as they show more disagreement among the annotators.} We repartition this data to simulate a multi-annotation setting, whether having \textbf{more than one label per training example} can be helpful. We reserve 500 randomly sampled examples for evaluation and use the rest for training.\footnote{The original datasets split data such that premise does not occur in both train and evaluation set. This random repartition breaks that assumption, now a premise can occur in both training and evaluation with different hypotheses. However, we find that the performance on examples with/without overlapping premise in the training set does not vary significantly.} % (see Table~\ref{tab:data}).
% \begin{table}
% \footnotesize
%     \centering
%     \begin{tabular}{l|r|r|r}
% \toprule
%  & \# train (single)& \# train (multi) & \# eval (multi)  \\ \midrule
% SNLI & 549K&1014 & 500\\ 
% MNLI & 392K&1099 & 500\\
%  \bottomrule
%     \end{tabular}
%     \caption{Data statistics. Multi-annotated examples are taken from ~\citet{Nie2020WhatCW} while single-annotated train set comes from the original SNLI and MNLI dataset.}
%     \label{tab:data}
% \end{table}

% All our experiments on all datasets use a batch size of 128. 

\subsection{Metrics}
Following ~\citet{Nie2020WhatCW}, we report classification accuracy, Jensen-Shannon Divergence~\cite{endres2003new}, Kullback-Leibler Divergence~\cite{kullback1951information}, comparing human label distributions with the softmax outputs of models. The accuracy is computed twice, once against aggregated gold labels in the original dataset (old), and against the aggregated label from 100-way annotated dataset (new). 
In addition, we report the model prediction label distribution entropy $H$.%(p)=-\sum_{i \in C}p_ilog(p_i)$,

\begin{table*} 
\footnotesize
\centering
 \begin{tabular}{l|l|l|r|r|r|r|r|r|r}
 \toprule
  & \multirow{2}{*}{\#annot}  &  \multicolumn{4}{c}{ChaosSNLI ($H=0.563$)} & \multicolumn{4}{|c}{ChaosMNLI ($H=0.732$)} \\ 
  & & JSD$\downarrow$ & KL$\downarrow$ &  acc (old/new)$\uparrow$ & $H$ & JSD$\downarrow$ & KL$\downarrow$ &  acc (old/new) $\uparrow$& $H$  \\ \midrule
RoBERTa && 0.250&0.547&	0.676 / 0.688&	0.363 & 0.312 & 0.753 & \textbf{0.628} / 0.578 & 0.444  \\
+ {pred smoothing} &150K  & 0.231&0.342&	0.676 / 0.688&	0.573 & 0.253 & 0.363 & \textbf{0.628} / 0.578 & 0.737\\
 % finetune & & 0.5K multi&\\ 
+ multi-annot   && \textbf{0.186}&\textbf{0.219}&	\textbf{0.684 / 0.732}&	0.643 & \textbf{0.195} & \textbf{0.183} & 0.616 / \textbf{0.684} & 0.910 \\ \midrule
RoBERTa  && 0.264&0.534&	\textbf{0.668} / 0.656&	0.445 & 0.319 & 0.686 &\textbf{0.552} / 0.496 & 0.518  \\
  + {pred smoothing} &15K  & \textbf{0.256}& 0.412 &	\textbf{0.668} / 0.656&	0.594 & 0.276 & 0.424 &\textbf{0.552} / 0.496 & 0.752\\
  + multi-annot && 0.293 &\textbf{0.336} &	0.624 / \textbf{0.674}&	0.985 & \textbf{0.252} & \textbf{0.269} & 0.546 / \textbf{0.554} & 1.026\\ 
 \bottomrule
  \end{tabular}
 \caption{Performances with smaller annotation budget. With smaller annotation budget, using multi-annotation can hurt accuracy on noisier evaluation setting (old), but still shows improvements on less noisy setting (new) and on most distribution metrics. }
    \label{tab:model_comparison}
\end{table*}

\subsection{Comparison Systems}
We use RoBERTa~\cite{Liu2019RoBERTaAR} based classification model, i.e., encoding concatenated hypothesis and premise and pass the resulting $[$CLS$]$ representation through a fully connected layer to predict the label distribution, trained with cross entropy loss.
\paragraph{Calibration Methods}
We experiment with three calibration methods~\cite{Guo2017OnCO,Miller1996AGO}. The first two methods are post-hoc and do not require re-training of the model. For all methods, we tuned a single scalar hyperparameter per dataset such that prediction label distribution entropy that matches that of human label distribution.

    \setlist{nolistsep}
\begin{itemize}[noitemsep]
\item \textbf{temp. scaling}: scaling by multiplying non-normalized logits by a scalar hyperparameter
\item \textbf{pred smoothing}: process softmaxed label distribution by moving $\alpha$ probability mass from the label with the highest mass to the all labels equally
\item \textbf{train smoothing}: process training label distribution by shifting $\alpha$ probability mass from the gold label to the all labels equally
\end{itemize}
%training label distribution smoothing (\textbf{train smoothing}),  , and post-hoc label smoothing. Label smoothing methods .% $\frac{\alpha}{|Y|-1}$.

\paragraph{Multi-Annotated Data Training}
We compare the results under the \textit{fixed} annotation budget, i.e., the number of annotations collected. We vary {number of examples annotated} and {the number of annotations per example}. We remove 10k randomly-sampled single annotated examples from training portion, and add 1k 10-way annotated examples from the re-partitioned training portion of the ChaosNLI dataset. For each example, we sample 10 out of 100 annotations. We first train model with single-annotated examples and further finetune it with multi-annotated examples.\footnote{We find merging multi-annotation data with single-annotation data does not show improvements.}
   %, 10-way annotated training examples and randomly exclude 1%, thus first train with single annotation data and then fine-tune with finely annotated data.

\subsection{Results}
Table~\ref{tab:main_result} compares the published results from ~\citet{Nie2020WhatCW} to our reimplementation and proposed approaches. As we set aside some 100-way annotated examples for training, our results are on the randomly selected subset of evaluation dataset, for sanity check, we report our re-implemented results on the full set, which matches the reported results.

The initial model was over confident, with smaller predicted label entropy (0.345/0.414) compared to the annotated label distribution entropy (0.563/0.732). We find all calibration methods improve performance on both distribution metrics (JSD and KL). Temperature scaling yields slightly better results than label smoothing, consistent with the findings from ~\citet{Desai2020CalibrationOP} which shows temperature scaling is better for in-domain calibration compared to label smoothing.% while label smoothing more effective for out-of-domain datasets. 
%with baseline (a.k.a. model trained on single-label data), baseline with calibration (temperature scaling and label smoothing) and model finetuned on densely-annotated data. 

Finetuning on multi-annotated data improves distribution metrics and the accuracy, yet the performance is below the estimated human performance. This approach seems to be effective at capturing the inherent ambiguity, and to learn label noise from the crowdsourced dataset. Our method shows more gains in accuracy on newly 100-way annotated dataset, with less noisy majority label, than the majority label from the original 5-way development set. We rerun the baseline and this model three times with different random seeds to determine the variance, which is small.\footnote{The standard deviation value of KL on all method / dataset pairs is lower than 0.01.} Using both multi-annotated data and calibration show mixed results, suggesting that calibration is not needed when you have multi-annotated examples. %the baseline is 0.009 and 0.004 on SNLI and MNLI, respectively. The \emph{std} of KL on our method is 0.003 on SNLI and 0.003 on MNLI.\emph{std}  The value of the std of both models are small.

%This shows that the finetuning method is able to better capture the inherent ambiguity of examples and potentially label noise. \footnote{We optimize the model and randomly pick 1K multi-annotated data under different random seeds.}

Table~\ref{tab:model_comparison} studies scenarios with smaller annotation budgets. Similar trend holds, yet in the smallest annotation budget setting, the gains in distribution metrics from fine-tuning approach comes at the cost of accuracy drop, potentially as the model is not exposed to diverse examples. % ( it improved significant gains on distribution metrics)

\begin{figure}[h]
\centering
\footnotesize
\setlength{\tabcolsep}{1.5pt}
\begin{tabular}{cc}
~~~(a) Human label entropy & (b) RoBERTa prediction entropy \\
% \raisebox{0.8em}{\rotatebox{90}{Frequency}}
\includegraphics[width=0.23\textwidth]{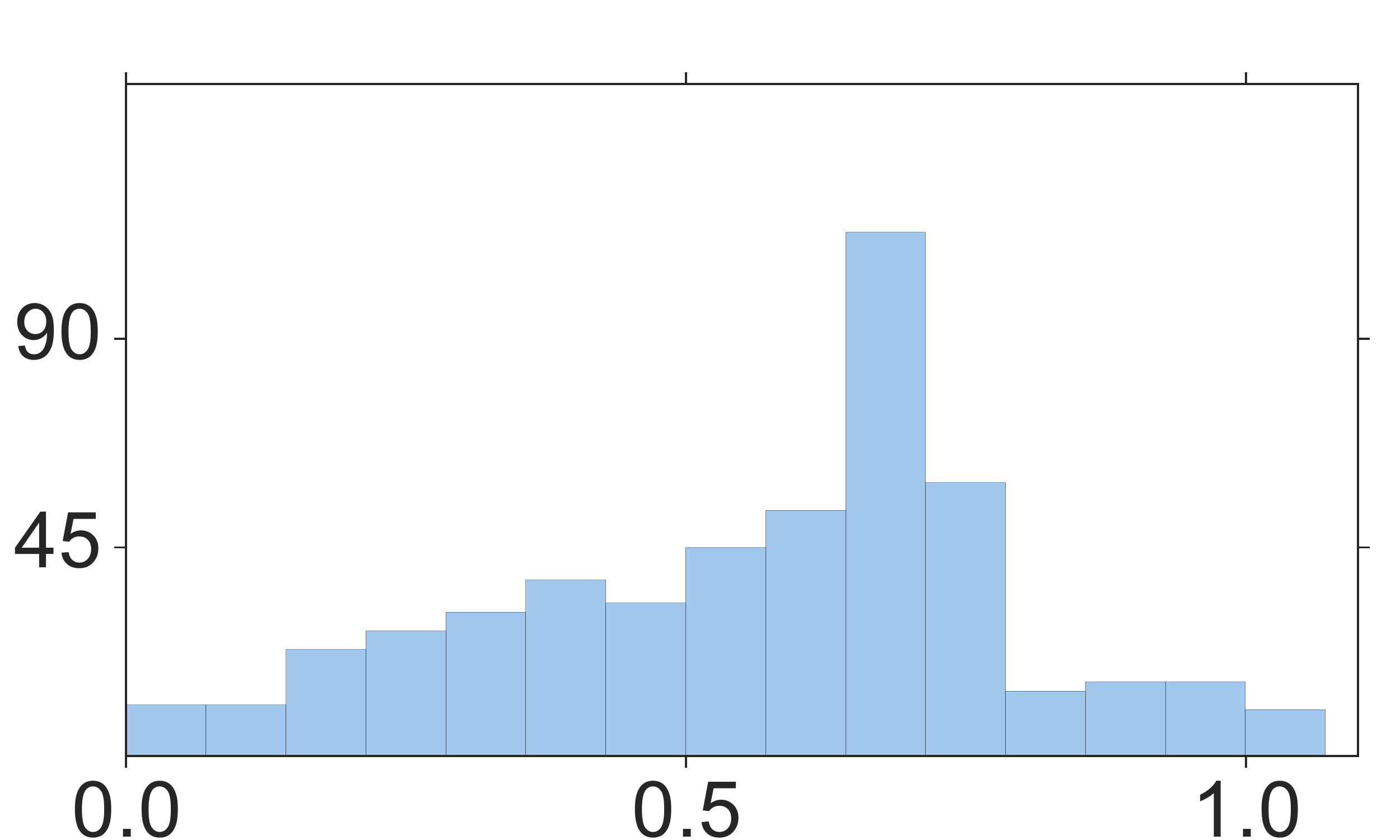}
& 
% \raisebox{1.2em}{\rotatebox{90}{\small }}
\includegraphics[width=0.23\textwidth]{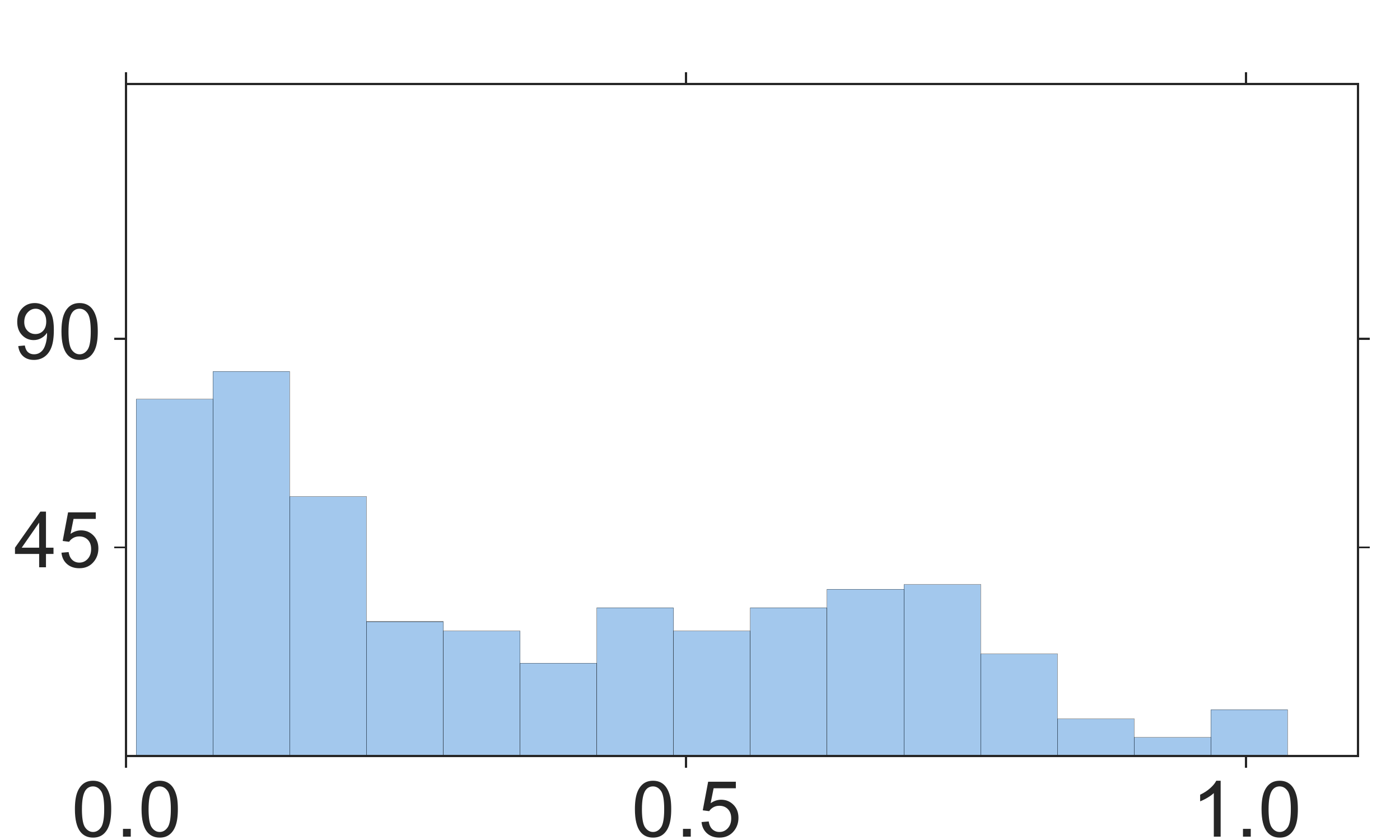}~~~~ \\
(c) Calibrated RoBERTa & (d) Multi-annot RoBERTa\\
% \raisebox{1.2em}{\rotatebox{90}{\small }}
\includegraphics[width=0.23\textwidth]{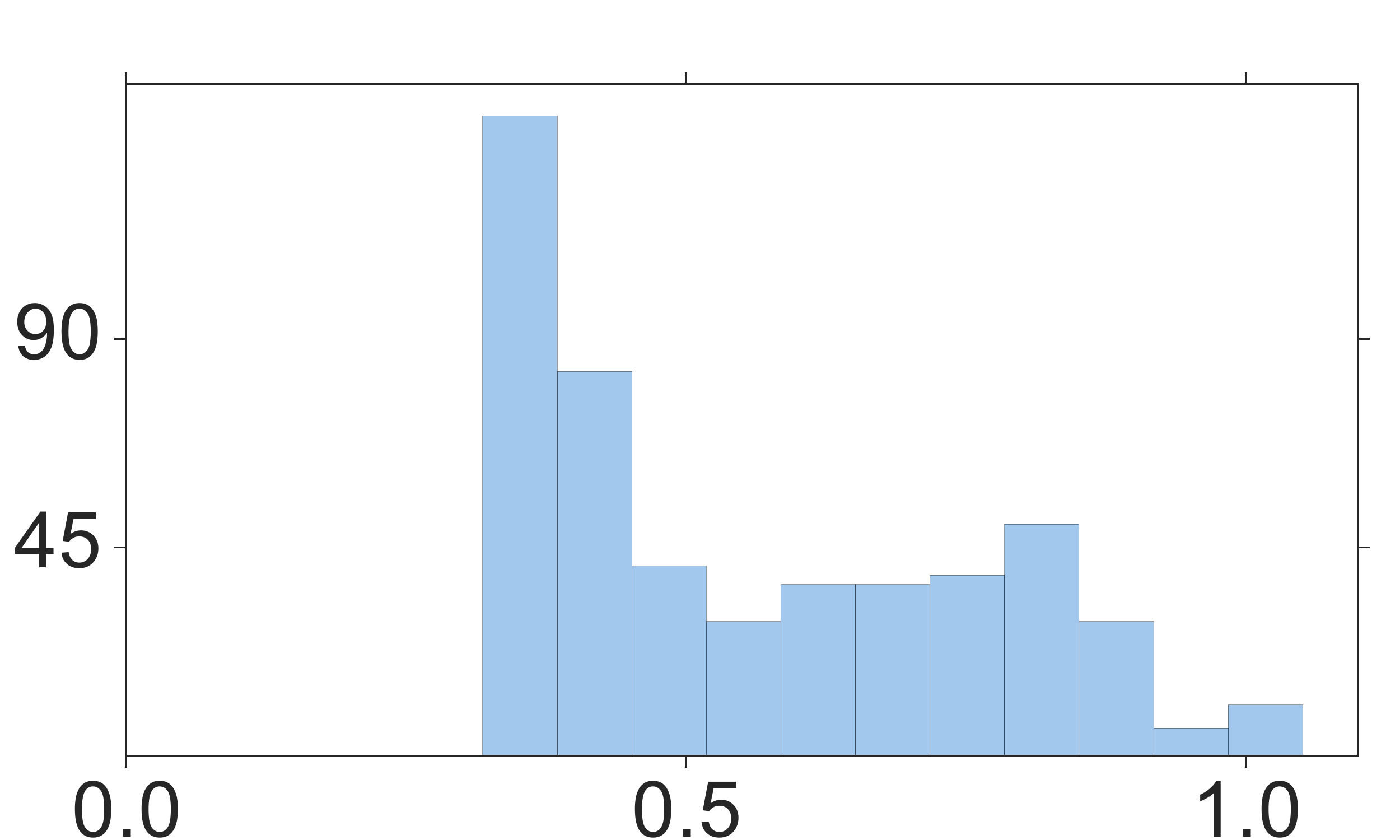} &
% \raisebox{1.2em}{\rotatebox{90}{\small }}
\includegraphics[width=0.23\textwidth]{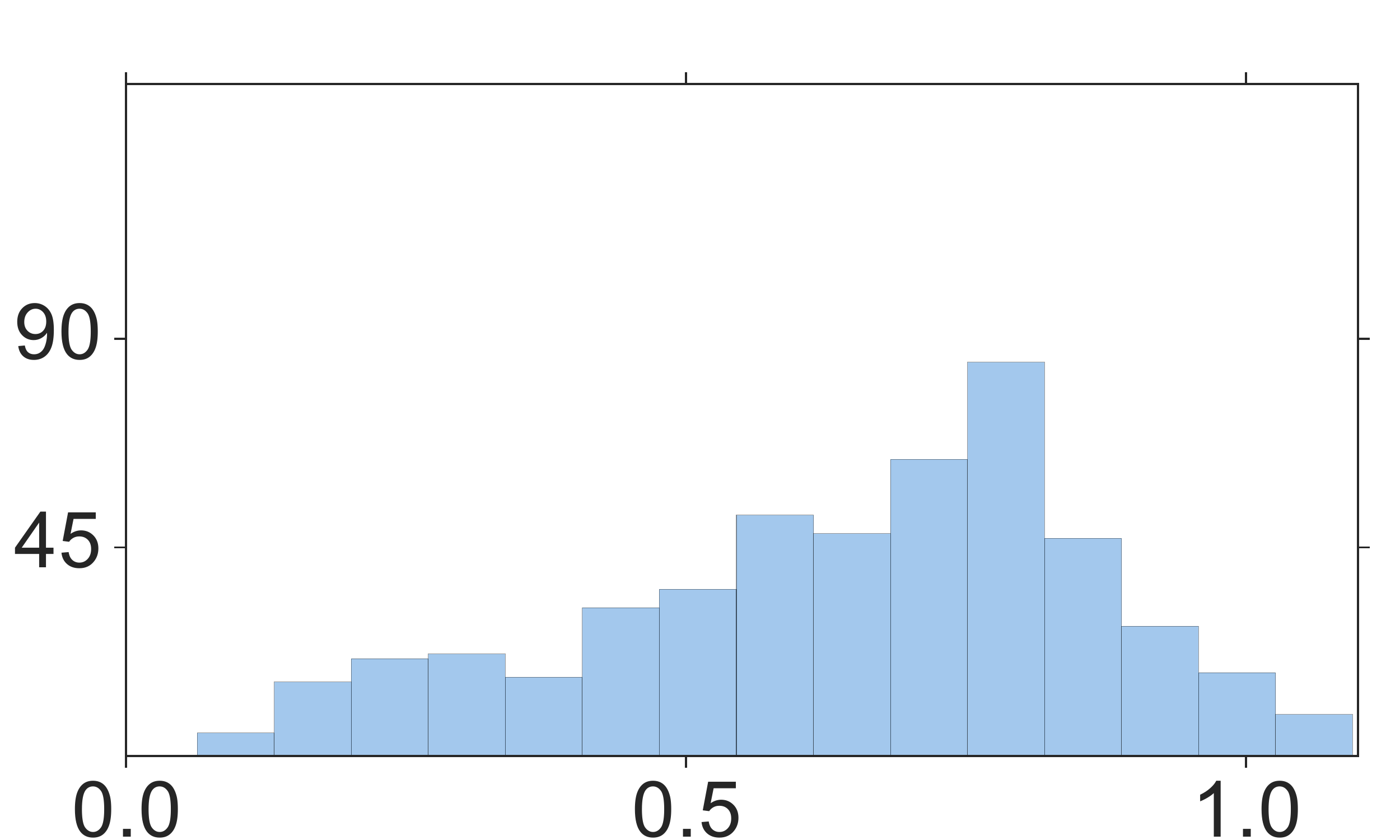}~~~~
\\
%\multicolumn{4}{c}{Entropy Value} \\
\end{tabular}
\caption{The empirical distribution of label/prediction entropy on ChaosSNLI dataset, where
x-axis denotes the entropy value and y-axis denotes the example count on the entropy bin.
Initial model prediction shows low entropy values for many examples, being over-confident. Post-hoc calibration successfully shifts the distribution to be less confident, but with artifacts of not being confident on any examples. Finetuning on the small amount of multi-annotated data (d) successfully simulate the entropy distribution of human labels.} 
\label{fig:labelentropy}
\end{figure}

\section{Analysis}
\textbf{Can we estimate the distribution of ambiguous and less ambiguous examples?} Figure~\ref{fig:labelentropy} shows the empirical example distribution over the entropy bins: The leftmost plot shows the annotated human label entropy over our evaluation set, and the plot next to it shows the prediction entropy of the baseline RoBERTa model predictions. Trained only on single label training examples, the model is over-confident about its prediction. With label smoothing, the over-confidence problem is relieved,  but still does not match the distribution of ground truth (see plot (c)). Training with multi-annotated data (plot (d)) makes the prediction distribution similar to the ground truth. 

\begin{figure}
\centering
\setlength{\tabcolsep}{1pt}
\begin{tabular}{ccccccc}
%\multicolumn{3}{c}{~~~\footnotesize{}}\\% & \multicolumn{3}{c}{\footnotesize{ChaosMNLI}} \\
%\raisebox{2em}{\rotatebox{90}{\footnotesize{JSD}}}
\includegraphics[width=0.13\textwidth,trim=10 6mm 10mm 6mm]{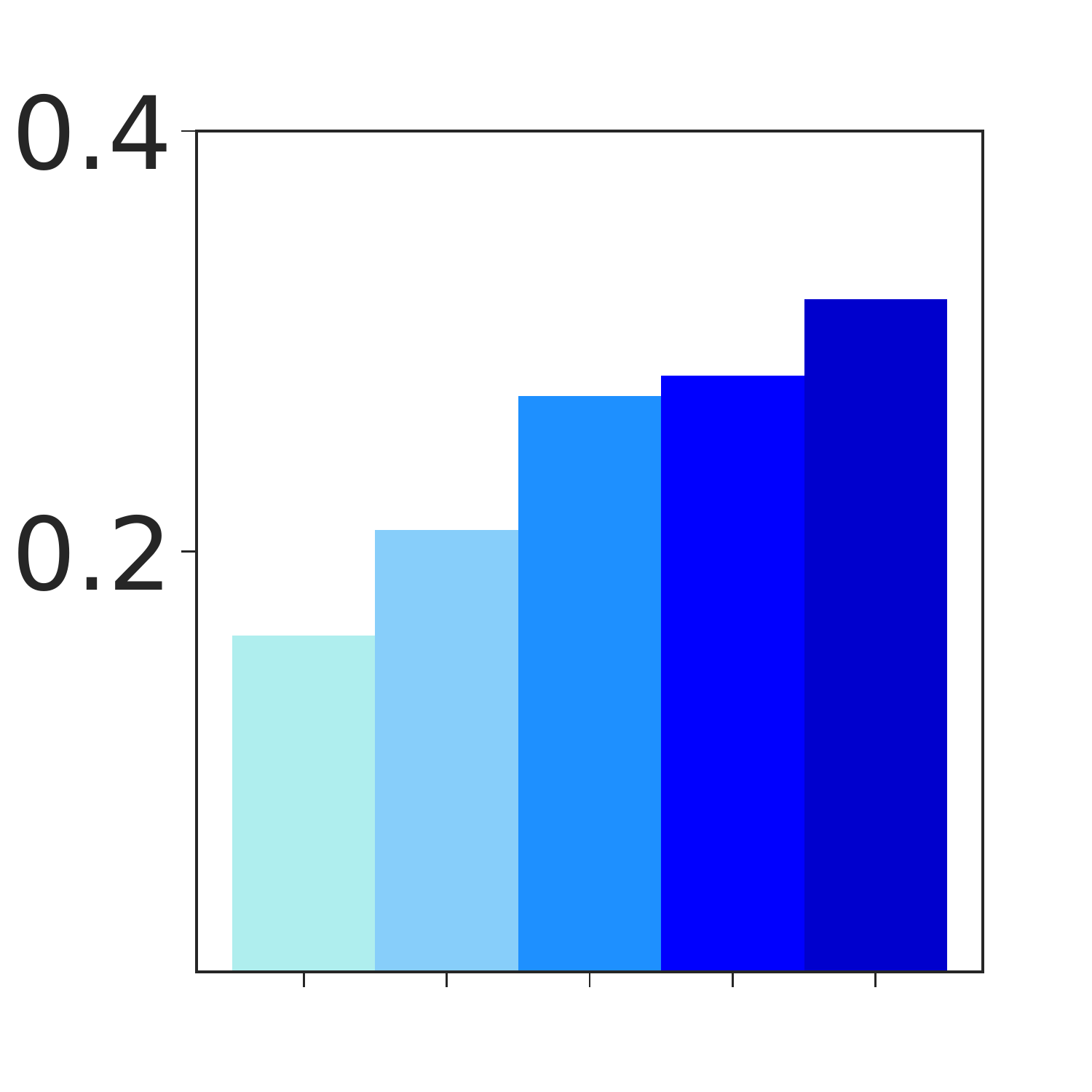}
& 
\includegraphics[width=0.13\textwidth,trim=10 6mm 10mm 6mm]{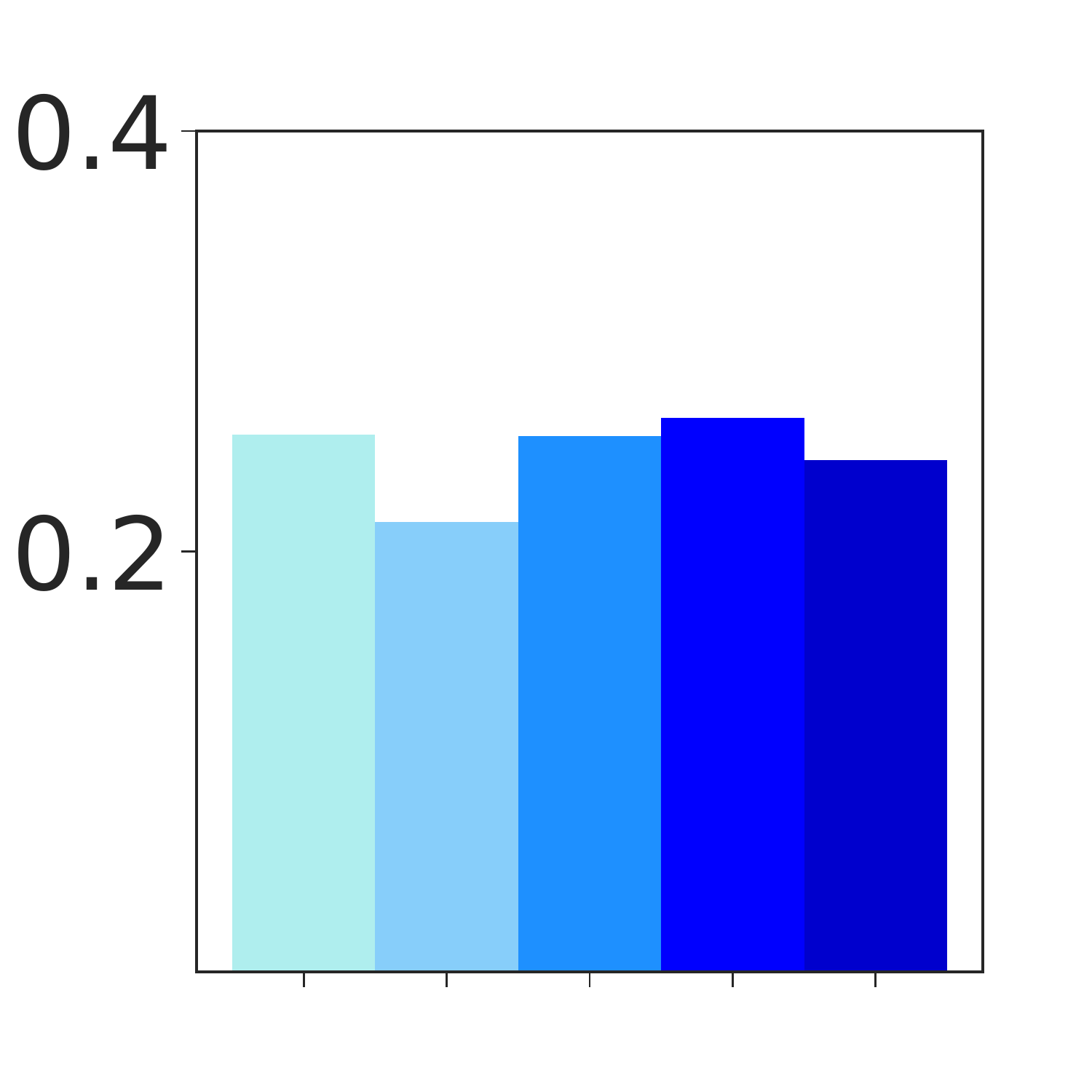} & 
\includegraphics[width=0.13\textwidth,trim=10 6mm 5mm 0mm]{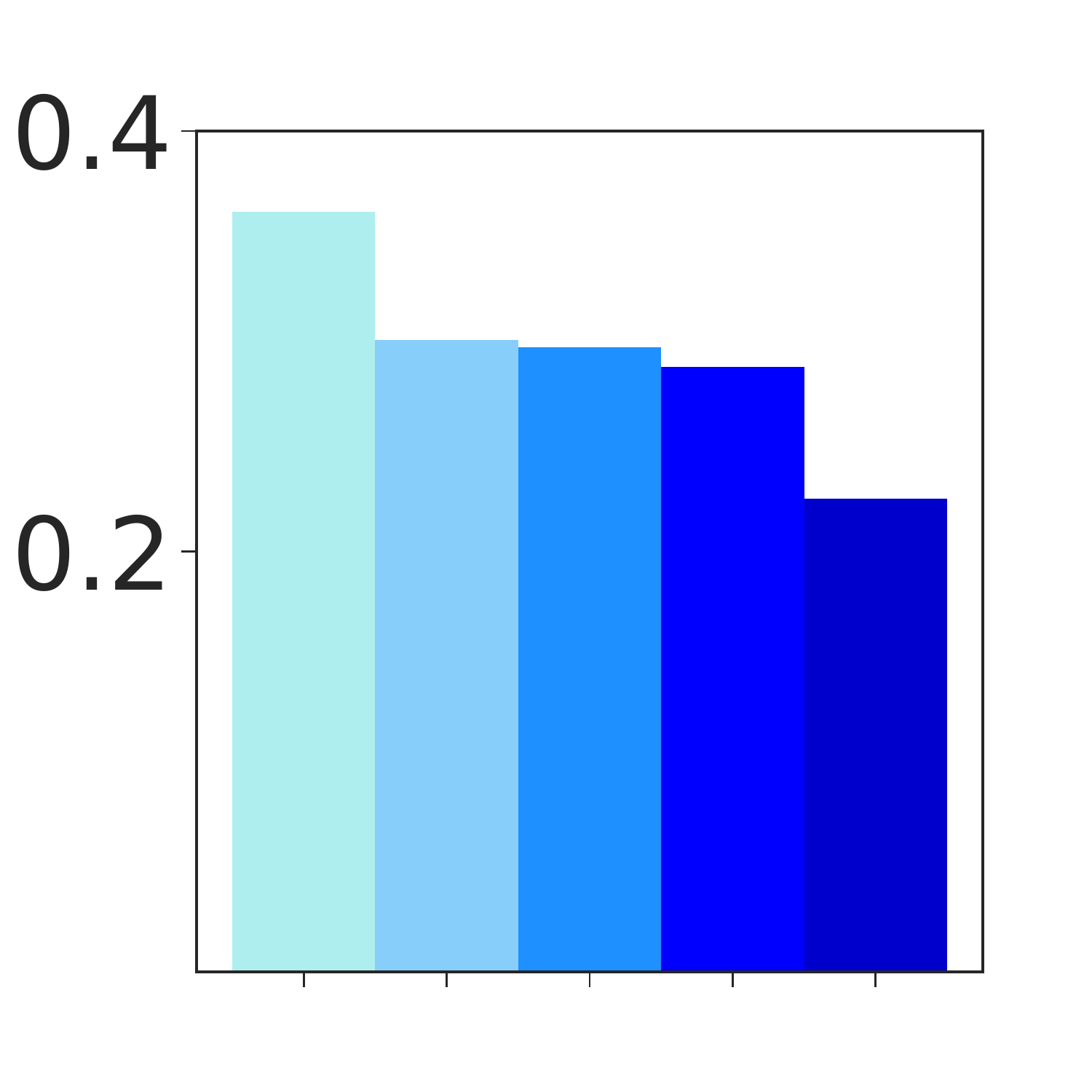} &
% \includegraphics[width=0.145\textwidth]{fig/jsd/mnli_alpha0.0.pdf}
% & [scale=0.55,trim=0 86mm 100mm 36mm]
% \includegraphics[width=0.145\textwidth]{fig/jsd/mnli_alpha0.3.pdf} & 
% \includegraphics[width=0.145\textwidth]{fig/jsd/mnli_alpha0.6.pdf} &
\hspace{-0.8em}
\raisebox{0.5em}{\includegraphics[width=0.1\textwidth]{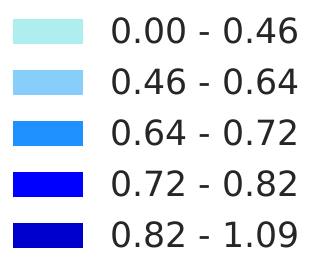}}

\\
\footnotesize{$\alpha$=0} &  \footnotesize{$\alpha$=0.3} & \footnotesize{$\alpha$=0.6 }\\%& \footnotesize{$\alpha$=0} &   \footnotesize{$\alpha$=0.3} &  \footnotesize{$\alpha$=0.6 }\\
\end{tabular}
\caption{Jensen-Shannon Divergence (JSD) on different label distribution entropy bins on ChaoSNLI dataset. Before label smoothing, the scores are lower for ambiguous examples, but the results swap after further smoothing.}
\label{fig:jsd_binned}
% \vspace{-10pt}
\end{figure}

\textbf{Are label distribution metrics reliable?} \citet{Nie2020WhatCW} suggests that ambiguous examples for humans are also more challenging for models in this dataset. We aware that this observation holds for accuracy measure but not for distribution metrics. 
Figure~\ref{fig:jsd_binned} demonstrates that when we further smooth the label distribution, JSD is better on supposedly more challenging examples where humans disagree. We find smoothing, which improves distribution metrics but not accuracy, can generate unlikely label distributions, e.g. assigning high values to both entailment and contradiction label. We hypothesize that humans often confuse between neutral and one of the two labels, not between these two labels. To quantify this intuition, we compute the average minimal probability assigned to either contradiction or entailment label: min value for human annotation is only 0.03, and 0.02 for the baseline model. We notice label smoothing increases min value to 0.06 and finetuning with multi-annotated increases the min value to 0.04.

We summarize studies not covered in our evaluation section (details in the appendix). 
\textbf{Do the results hold for other model architecture or bigger model?} Yes. Our results show the same pattern with ALBERT model~\cite{Lan2020ALBERTAL} and the larger variant of RoBERTa model.
\textbf{Is the method sensitive to the number of labels per training example?} No. We try different label strategies (5-way, 10-way, or 20-way), while keeping the total number of annotation fixed, and observe no changes. % Using  annotated training instances showed similar gains. %Introducing \textit{some} disagreement into training seems to be sufficient. 
\textbf{What happens if you keep smoothing the distribution?} We choose label smoothing hyper-parameter such that the predicted label distribution entropy matches that of human annotation distribution. However, we notice keep smoothing model prediction further ($\alpha =0.125 \rightarrow 0.4$ for SNLI) brings further gains.
\textbf{Should we carefully select which examples to have multiple annotations?} Maybe. We experiment on how to select examples to have multiple annotations, using the ideas from ~\citet{Swayamdipta2020DatasetCM}. We finetune with 100 most hard-to-learn, most easy-to-learn, most ambiguous, and randomly sample examples from 1K examples. Easy-to-learn examples, with lowest label distribution entropy, are the least effective, but the difference is small in our settings. 

% the entropy plot shifts to the right (c), but still does not match the label distribution. When we fine tune the model on densely annotated data, the entropy distribution (d) now resembles the label distribution that we aims to achieve.

% Figure~\ref{fig:labelentropy} empirically shows that, compared to ground-truth label entropy, the model trained with single-annotated data is over-confident about its prediction. The entropy of the prediction is much smaller than ground-truth label entropy.
% By using label smoothing,) shows that the uncertainty is still far from the ground-truth. 

%, when we label smooth the predictions, JSD 
%that 1) using label smoothing cannot always improve the JSD score and 2) increasing $\alpha$ smoothly changes the distribution. Increasing $\alpha$ from 0 to 0.6, the JSD value first decreases and then increases.
% It indicates that $\alpha$ is sensitive and cannot truly recover the human label distribution.

%% file: tex/related.tex
\section{Related Work}
% \vspace{-4pt}
\paragraph{Human Disagreement in NLP}
Prior works have covered inherent ambiguity in different language interpretation tasks. \citet{Aroyo2015TruthIA} demonstrates that it is improper to believe that there is a single truth for crowdsourcing. Question answering, summarization and translation literatures have been collecting multiple references per example for evaluation. Most related to our work, ~\citet{Pavlick2019InherentDI} carefully examines the distribution behind human references and ~\citet{Nie2020WhatCW} has conducted a larger-scale data collection. To capture the subtleties of NLI task, ~\citet{Glickman2005APC,Zhang2017OrdinalCI,Chen2020UncertainNL} introduce graded human responses on the subjective likelihood of an inference. In this work, we explicitly focus on improving label distribution estimation on the existing benchmarks. % instead of three-way classification. 

\paragraph{Efficient Labeling} Previous works have explored different data labeling strategies for NLP tasks, from active learning~\cite{Fang2017LearningHT}, providing fine-grained rationales~\cite{Dua2020BenefitsOI} to model prediction validation approaches~\cite{kratzwald2020learning}. Recent work~\cite{mishra2020we} studies how much annotation is necessary to solve NLI task. In this work, we study collecting multiple references per each training example, which has not been explored to our understanding.

\paragraph{Calibration} in NLP~\cite{Nguyen2015PosteriorCA,Ott2018AnalyzingUI} could make predictions more useful and interpretable. Large-scale pretrained models are not well calibrated~\cite{Jiang2018ToTO}, and its predictions tend to be over-confident~\cite{Malkin2009MultilayerRS,Thulasidasan2019OnMT}. Calibration has been studied for pretrained language models for classification~\cite{Desai2020CalibrationOP}, reading comprehension~\cite{kamath2020selective} and in general machine learning topics \citep[e.g.][]{Guo2017OnCO, pleiss2017fairness}. While these works focus on improving robustness to out-of-domain distribution, we study \textit{predicting} label distributions.

%% file: tex/conclusion.tex
\section{Conclusion}
We study capturing inherent human disagreement in the NLI task through calibration and using a small amount of multi-annotated training examples. Annotating fewer examples many times as apposed to annotating as many examples as possible can be useful for other language understanding tasks with ambiguity and generation tasks where multiple references are valid and desirable~\cite{Hashimoto2019UnifyingHA}. 

%% file: tex/appendix.tex
\section*{Appendix}

\subsection*{Hyperparameters And Experimental Settings}
Our implementation is based on the \textit{HuggingFace Transformers}~\citep{wolf2020transformers}.
We optimize the KL divergence as objective with the Adam optimizer \citep{kingma2014adam} and batch size is set to 128 for all experiments.
The Roberta-base and Albert are trained for 3 epochs on single-annotated data.
For the finetuning phase, the model is trained for another 9 epochs. 
The learning rate, $10^{-5}$,  is chosen from AllenTune \citep{dodge2019show}.
For post-hoc label smoothing, 
we try $\alpha \in \{0.1, 0.125, 0.15$ $, ..., 0.6\}$ and choose $\alpha^*$ such that predicted label entropy matches the human label distribution entropy. For SNLI, the value is 0.125, and 0.225 for MNLI.
For post-hoc temperature scaling, we try temperature in $\{1.5, 1.75, 2, ..., 5\}$ and choose the temperature such that predicted label entropy matches the human label distribution entropy (1.75 for SNLI and 2 for MNLI).
For training label smoothing, we follow the post-hoc label smoothing value.

\subsection*{Additional Experiments}
Table~\ref{tab:label_count_comparison} shows ablations on different label collection strategies. While keeping the total number of annotations, we change the number of multi-annotated data and the number of annotation per multi-annotated example. We find that performance improvement do not vary significantly across the varying settings.

\begin{table}[h]
\centering
% \scriptsize
    \centering
    \begin{tabular}{l|l|r|r|r|r}
\toprule
  %& & \multicolumn{4}{c}{SNLI} & \multicolumn{4}{|c|}{MNLI} & \multicolumn{4}{c}{NLI$\alpha$}  \\ 
\# multi  & \# single&JSD & KL &  acc (old/new) & $H$ \\ \midrule
0 &150K& 0.25&0.55&	0.676 / 0.688&	0.363 \\
  0.5K (20-way)  & 130K & 0.20&0.22&	0.676 / 0.726&	0.695\\
 % 0.5K (10-way)& 145K & 0.19&0.22&	0.684 / 0.736&	0.694 \\ 
  1K (10-way) &140K& 0.19&0.22&	0.684 / 0.732&	0.643\\ 
  5K (5-way) &145K& 0.19&0.22&	0.676 / 0.732&	0.701 \\ 
 \bottomrule
    \end{tabular}
    \caption{Label count comparison on SNLI dataset. The total number of annotation is consistent among different rows. }
    \label{tab:label_count_comparison}
\end{table}

\begin{figure}[h]
% \footnotesize
\centering
\setlength{\tabcolsep}{3pt}
\begin{tabular}{cc}
~~~(a) SNLI & (b) MNLI\\
\raisebox{1.0em}{\rotatebox{90}{KL divergence}}
\includegraphics[width=0.23\textwidth]{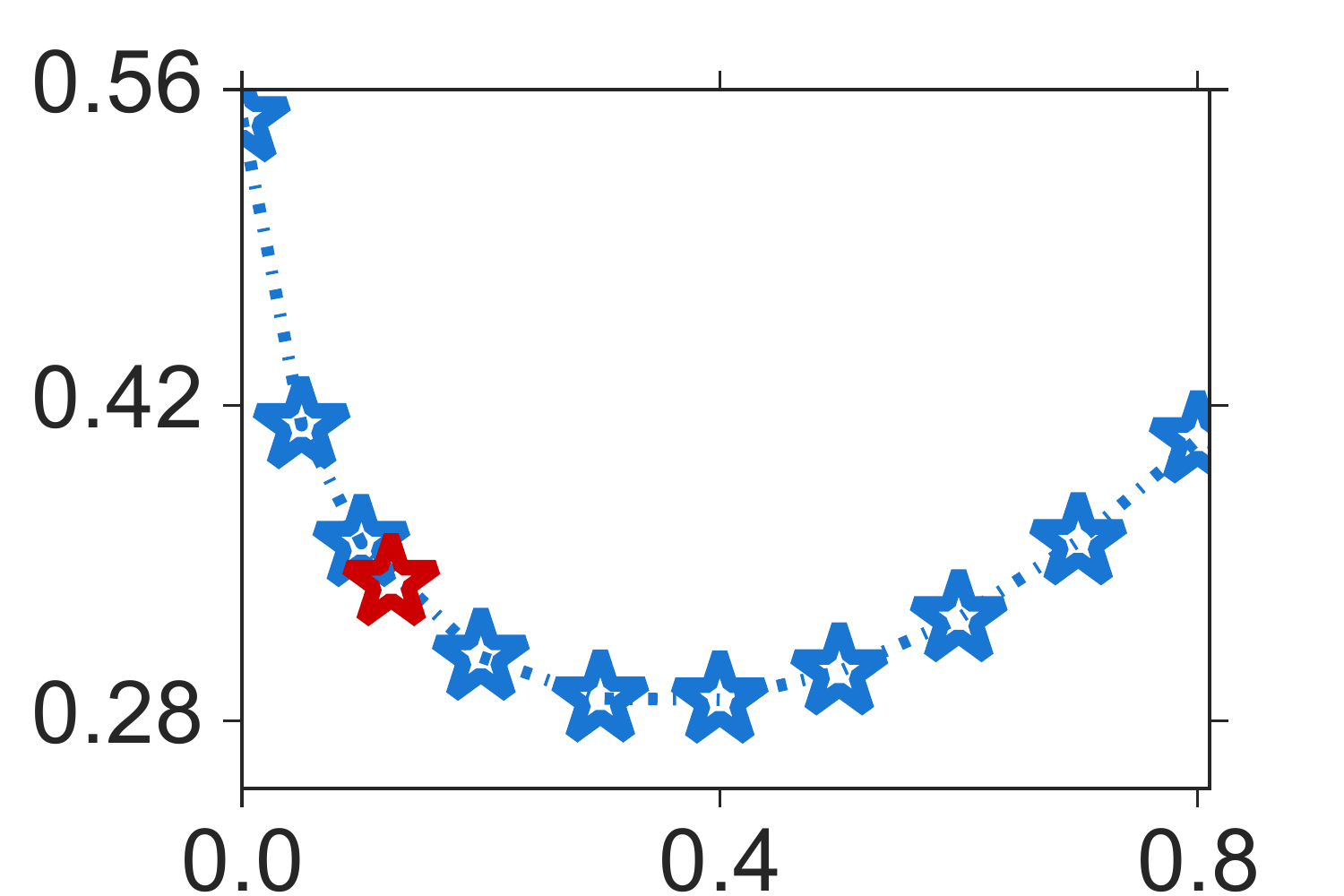}
& 
\raisebox{1.2em}{\rotatebox{90}{\small }}
\includegraphics[width=0.23\textwidth]{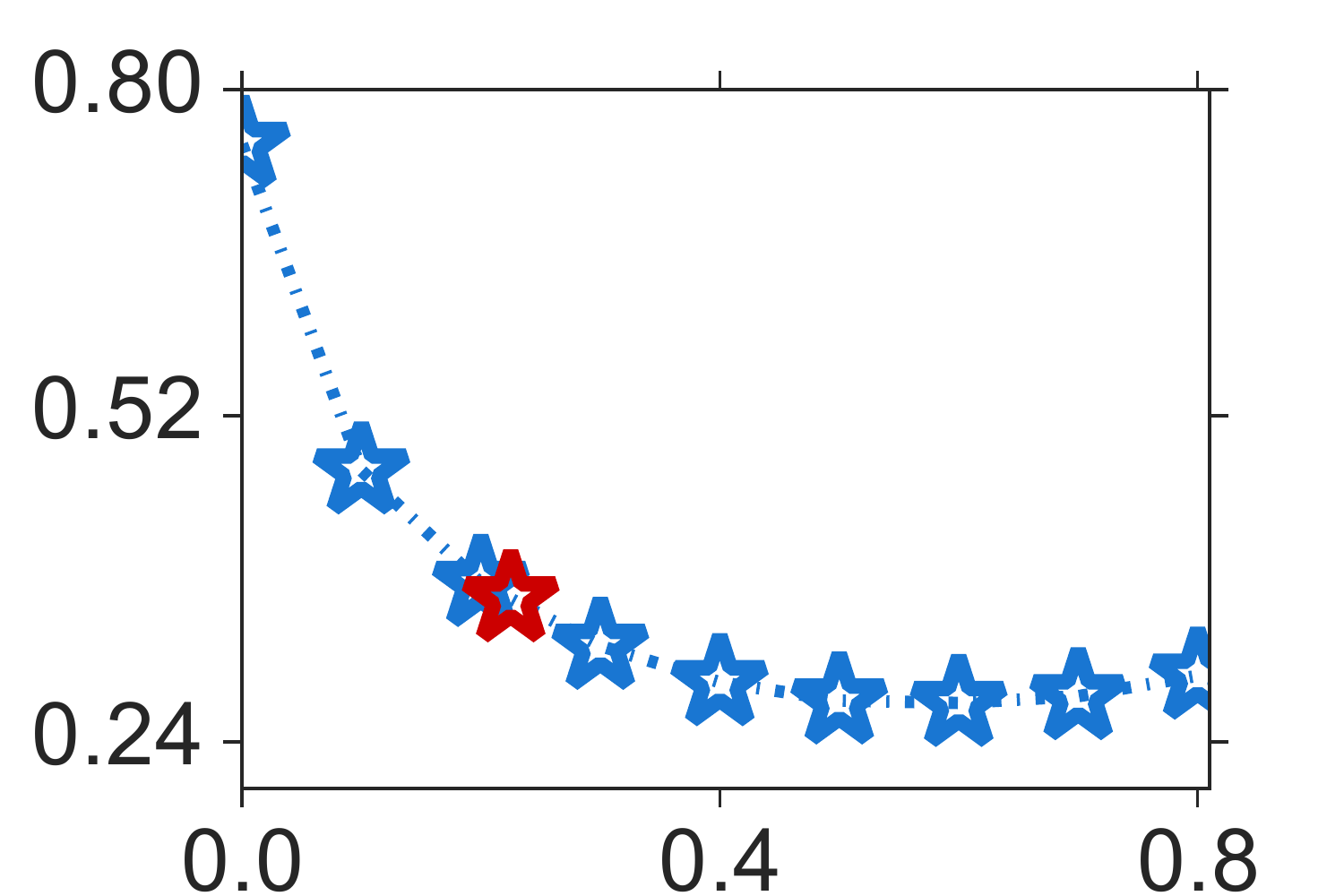}
\\
\multicolumn{2}{c}{Label Smoothing $\alpha$} \\
\end{tabular}
\caption{The KL divergence with different value of label smoothing $\alpha$. The \textcolor{red}{red} star refers to the case when the prediction entropy is the same as the label entropy, and \textcolor{blue}{blue} stars denote the data points.} 
\label{fig:labelsmoothing_kl}
\end{figure}

Figure~\ref{fig:labelsmoothing_kl} shows how the label smoothing hyperparameter $\alpha$ impacts the KL divergence score. When increasing $\alpha$, the KL divergence first decreases and then increases.

Table \ref{tab:model_comparison_appendix} shows that, with a different model (ALBERT), we can still have consistent result and conclusion.

\begin{table*}[h]
 \footnotesize
\centering
 \begin{tabular}{l|l|l|r|r|r|r|r|r|r}
 \toprule
  & \multirow{2}{*}{\#annot}  &  \multicolumn{4}{c}{ChaosSNLI ($H=0.563$)} & \multicolumn{4}{|c}{ChaosMNLI ($H=0.732$)} \\ 
  & & JSD & KL &  acc (old/new) & $H$ & JSD & KL &  acc (old/new) & $H$  \\ \midrule
ALBERT &&  0.243& 0.474&	0.668 / 0.684&	0.422 & 0.314 & 0.735 & \textbf{0.596} / 0.544 & 0.496 \\
 + post-hoc smoothing &150K & 0.231&0.344&	0.668 / 0.684&	0.581 & 0.268 & 0.410 & \textbf{0.596} / 0.544 & 0.742\\
+ multi-annot && \textbf{0.201}&\textbf{0.225}&	\textbf{0.676} / \textbf{0.720}&	0.709 & \textbf{0.217} & \textbf{0.218} & 0.584 / \textbf{0.634} & 0.948  \\
 \bottomrule
  \end{tabular}
 \caption{Ablation studies: performance with a different model (ALBERT). }
    \label{tab:model_comparison_appendix}
\end{table*}

\subsection*{Examples}
We present examples where our model improved baseline predictions in Table~\ref{tab:showcase}. Original model produces over-confident predictions (row 1, 2 and 5).
The over-confidence predictions are fixed with further finetuning with multi-annotated data. On a few examples, finetuning further improves the accuracy. 

\begin{table*}
\centering
\scriptsize
\begin{tabular}{p{16em}p{16em}ccc}
\toprule
\multirow{2}{*}{\bf Premise} &  \multirow{2}{*}{\bf Hypothesis} &  {\bf Human Label Dist} & {\bf Baseline} & {\bf Finetuned}\\
 & & \multicolumn{3}{c}{[Entailment, Neutral, Contradiction]} \\
\midrule
Six men, all wearing identifying number plaques, are participating in an outdoor race. & A group of marathoners run. & [0.31, 0.68, 0.01] & [0.01, 0.97, 0.02] & [0.27, 0.65, 0.08] \\
\midrule
A small dog wearing a denim miniskirt. & A dog is having all its hair shaved off. & [0.00, 0.40, 0.60] & [0.00, 0.03, 0.97] & [0.03, 0.29, 0.68] \\
\midrule
A goalie is watching the action during a soccer game. & The goalie is sitting in the highest bench of the stadium. & [0.01, 0.84, 0.15] & [0.00, 0.39, 0.61] & [0.02, 0.81, 0.17]\\
\midrule
Two male police officers on patrol, wearing the normal gear and bright green reflective shirts. & The officers have shot an unarmed black man and will not go to prison for it. & [0.00, 0.66, 0.34] & [0.00, 0.15, 0.85] & [0.01, 0.59, 0.40] \\
\midrule 
An african american runs with a basketball as a caucasion tries to take the ball from him. & The white basketball player tries to get the ball from the black player, but he cannot get it. & [0.37, 0.61, 0.02] & [0.04, 0.92, 0.04] &  [0.23, 0.70, 0.07]\\\midrule
A group of people standing in front of a club. & There are seven people leaving the club. & [0.02, 0.78, 0.20] & [0.02, 0.34, 0.64] & [0.02, 0.66, 0.32] \\
\bottomrule
\end{tabular}
\caption{Examples from ChaosSNLI development set. `Baseline' and `Finetuned' denote the model trained on single-annotation data and the model further finetuned on multi-annotated data, respectively.}
\label{tab:showcase}
\end{table*}